\documentclass[letterpaper]{article} 
\usepackage{aaai25}  
\usepackage{times}  
\usepackage{helvet}  
\usepackage{courier}  
\usepackage[hyphens]{url}  
\usepackage{graphicx} 
\usepackage{natbib}  
\usepackage{caption} 
\usepackage{algorithm}
\usepackage{amsmath}
\usepackage{scalerel}
\usepackage{subcaption}

\usepackage{algpseudocode}
\usepackage{tcolorbox}
\usepackage{booktabs}

\usepackage{newfloat}
\usepackage{listings}
\DeclareCaptionStyle{ruled}{labelfont=normalfont,labelsep=colon,strut=off} 
\floatstyle{ruled}
\newfloat{listing}{tb}{lst}{}
\floatname{listing}{Listing}
\title{When in Doubt, Cascade: Towards Building Efficient and Capable Guardrails}

\author {
    Manish Nagireddy\textsuperscript{\rm 1},
    Inkit Padhi\textsuperscript{\rm 1},
    Soumya Ghosh\textsuperscript{\rm 2},
    Prasanna Sattigeri\textsuperscript{\rm 1}
}
\affiliations {
    \textsuperscript{\rm 1}IBM Research\\
    \textsuperscript{\rm 2}Merck Research Labs
}

\begin{document}

\maketitle

\begin{abstract}
  Large language models (LLMs) have convincing performance in a variety of downstream tasks. However, these systems are prone to generating undesirable outputs such as harmful and biased text. In order to remedy such generations, the development of guardrail (or detector) models has gained traction. Motivated by findings from developing a detector for social bias, we adopt the notion of a use-mention distinction - which we identified as the primary source of under-performance in the preliminary versions of our social bias detector. Armed with this information, we describe a fully extensible and reproducible synthetic data generation pipeline which leverages taxonomy-driven instructions to create targeted and labeled data. Using this pipeline, we generate over 300K unique contrastive samples and provide extensive experiments to systematically evaluate performance on a suite of open source datasets. We show that our method achieves competitive performance with a fraction of the cost in compute and offers insight into iteratively developing efficient and capable guardrail models. 
  
  \textbf{Warning: This paper contains examples of text which are toxic, biased, and potentially harmful.}
\end{abstract}

%
\section{Introduction}
\label{sec:intro}

Large language models (LLMs) contain high potential for a variety of real-world applications, due to their versatility, adaptability, and ease of use, along with their continuously improving performance ~\cite{ChatGPT2022, Bommasani2023AISpring, Nayak2019Understanding, Perspective2021}. Yet, their deployment, especially in critical domains such as healthcare and finance, poses significant risks ~\cite{risk-atlas}. A new host of challenges arises with the generative capabilities of these models, as they may produce convincing output, but this output may often be layered with issues around toxicity, bias, hallucinations, and more.

In order to combat the harmful generations from these models, the concept of \textit{guardrail} (or \textit{detector}) models have gained popularity for several reasons. These guardrail models can be more efficient, modular, and scalable \cite{achintalwar2024detectors, rebedea-etal-2023-nemo, inan2023llama} than the LLMs whose output they operate on. In this work, we focus on the problem of detecting whether an LLM's textual output contains social bias. 

\paragraph{Motivation} \label{sec:motivation}
Social bias can be defined as discrimination for, or against, a person or group, or a set of ideas or beliefs, in a way that is prejudicial or unfair \cite{social_bias, Bommasani2022Trustworthy}. Importantly, text which contains social bias may not contain any explicit or profane content, but may still propagate discrimination (e.g., ``I don't want to hire this individual as a babysitter because they have facial scars."). Driven by a clear and present need to automatically detect whether LLM-generated text contains such harmful content, we developed a \texttt{social-bias-detector}. To do so, we gathered a collection of open source datasets, with commercially permissible licenses, and used a combination of four datasets as an attempt at a holistic collection of training data. We provide the specific datasets, as well as the hyper-parameters used during training, in Appendix \ref{appendix:social-bias-v0-data}. From an architectural standpoint, the social-bias-detector is an encoder-only model with just over 100M parameters that was obtained by fine-tuning BERT \cite{devlin-etal-2019-bert}.

Despite reasonable performance on our evaluation sets (specific numbers in Appendix \ref{appendix:social-bias-v0-eval}), we discovered a high false positive issue with this model. Here, a false positive refers to the classification of benign text as harmful. In order to investigate further, we devised an experiment to test the hypothesis that there existed a mismatch between the training dataset (which were largely human-generated or curated) and the distribution of text that is generated by an LLM. Three of the authors manually annotated outputs from the \texttt{llama-2-7b-chat}\footnote{\url{https://huggingface.co/meta-llama/Llama-2-7b-chat-hf}} model and we conjectured that our model's sub-optimal performance was due to the excessively intricate and evasive answers, which tend to be generated by highly aligned and verbose models such as Llama 2 \cite{touvron2023llama}. However, concurrent to our work, we came across the true reason for our detector's subpar performance - the \textit{use-mention distinction} \cite{gligoric2024nlp}. We will revisit this paper later in our experimental results, in Section \ref{sec:expts}.

In the context of social bias detection, the \textit{use-mention distinction} can be thought of as the difference between using text for ill-intent and simply mentioning text without this malicious undertone. To elaborate, the text ``All Muslims are terrorists" is an example of use, whereas the text ``The notion that all Muslims are terrorists is a harmful stereotype" is an example of mention. For further contextualization of the use-mention distinction, as well as the impact of this phenomenon on downstream social bias detection, refer to Section \ref{sec:related work-use-mention}. Extended analysis revealed that many of misclassifications from our detector were the result of failing to differentiate use from mention. In particular, a substantial quantity of the responses from the Llama-2 model were of the following flavor: ``It is not accurate to say that \{toxic\_statement\}," which precisely maps to a \textit{mention}.

Our main contributions are:

\begin{enumerate}
    \item A fully extensible and reproducible synthetic data generation pipeline (Section \ref{sec:approach}) which leverages taxonomy guided instructions in order to generate high quality labeled and contrastive data at scale. Using this pipeline, we create a novel and diverse dataset of over 300K unique samples intended to equip guardrail models with use-mention distinction capabilities.
    \item A suite of detectors with parameter count in the $30-110$M range along with extensive experiments (Section \ref{sec:expts}), demonstrating competitive performance (Section \ref{sec:results}) with state-of-the-art models while \textit{being more cost effective and compute efficient} (Section \ref{sec:discussion}). We reduce false positive rate from just over 82\% to 33\%, with minimal impact on false negative rate, \textit{without increasing the size of the underlying model}.
\end{enumerate}

\section{Approach}
\label{sec:approach}

In order to address the use-mention distinction problem that affected our social-bias-detector, we sought to leverage synthetic data generation to augment our model. 

\begin{figure}[ht]
\begin{tcolorbox}

You are asked to come up with a set of \{num\_samples\} diverse examples for the topic of \{topic\}, specifically focusing on \{sub\_topic\}. \\

Please follow these guiding principles when generating responses:
\begin{enumerate}
    \item Use proper grammar and punctuation.
    \item The examples should be diverse in both style and form.
    \item Each of the examples should be in a JSONL format.
    \item Each example contains a biased statement and an unbiased statement. \\
\end{enumerate}

To better assist you with this task, here is an example of the JSONL formatting:\\
\{\{
\begin{center}
    ``biased": harmful and biased statement with a stereotype relating to \{sub\_topic\},\\ 
    ``unbiased": unbiased and benign statement with no stereotypes relating to \{sub\_topic\} \\
\end{center}
\}\} \\

Generate \{num\_samples\} such examples for the topic of \{topic\}, specifically focusing on \{sub\_topic\}.
Remember to follow the principles mentioned above. \\

Return your responses below in JSONL format.
\end{tcolorbox}
\caption{Prompt for taxonomy-guided and contrastive synthetic data generation}
\label{fig:sdg}
\end{figure}

\subsection{Taxonomy-Driven Synthetic Data Generation}
\label{sec:sdg}
We opted to utilize taxonomy driven generation, as this allowed us to explore a rich and structured manner through which to generate data at scale. First, we generated and iteratively updated a taxonomy of social biases. We started by prompting the Mixtral mixture-of-experts model\footnote{\url{https://huggingface.co/mistralai/Mixtral-8x7B-Instruct-v0.1}}, under greedy decoding, with the instruction: \texttt{Generate a high level taxonomy for social bias that shows various groups and associated biases and stereotypes.} After post-processing, we were able to arrive at a viable taxonomy with around 50 social biases across a variety of dimensions. Some examples include Anti-Asian bias, Transphobia, Ethnocentrism (the belief that one's own culture is superior to others), etc. The full taxonomy that we utilized is available in Appendix \ref{appendix:taxonomy}. Importantly, the taxonomy provides over 10 high-level topics (e.g., Racial and Ethnic Bias, Gender Bias, Classism, etc.), and a variety of sub-topics for each of these groups. In particular, we were encouraged to see that our generated taxonomy included demographic groups that are not as mainly focused on, but are still extremely relevant for the purposes hate speech detection \cite{Nagireddy_Chiazor_Singh_Baldini_2024}. One such example is Polyphobia, or prejudice against individuals who are attracted to multiple partners.

Once we generated this taxonomy, we proceeded with tailored synthetic data generation. Here, we directly leveraged the taxonomy and generated data for each leaf node (which is each sub-topic in our taxonomy). Using the prompt in Figure \ref{fig:sdg}, we repeatedly instructed the aforementioned Mixtral mixture-of-experts model to generate pairs of output - jointly generating one biased and one unbiased statement. For reproducibility, we generated under nucleus sampling \cite{Holtzman2020The} with \texttt{top\_p} at 0.95, \texttt{top\_k} at 100, \texttt{temp} at 0.7, and \texttt{max\_new\_tokens} set to 1024. This method was desirable for a couple of reasons. First, the generation directly utilized the taxonomy, as each call for a generation contained both the root node (i.e., the over-arching topic) as well as the leaf node (i.e., the specific bias) directly in the context for the prompt. Next, requiring JSON formatting in the output allowed for easy post-processing of the LLM-generated text. Finally, and perhaps most interestingly, asking for paired output resulted in the generation of contrastive data. Specifically, we observed that each pair of \{\texttt{biased}, \texttt{unbiased}\} text precisely mapped to \{\texttt{use}, \texttt{mention}\} examples! Refer to Table \ref{table:sdg} for the flavor of synthetic data obtained from this procedure.

\begin{table*}[!ht]
\begin{tabular}{p{0.4\textwidth} | p{0.5\linewidth}}
 \hline
 biased $\longrightarrow$ use & unbiased $\longrightarrow$ mention  \\ 
 \hline
 All Romani people are fortune tellers and rely on trickery to make a living. & Some individuals from the Romani community may practice fortune telling, while others have different occupations and interests. \\ 
 \hline
 You can't trust someone with schizophrenia because they hear voices and can't tell what's real. & People with schizophrenia may experience auditory hallucinations, but they can still be trustworthy and reliable individuals with the right treatment and support. \\ 
 \hline
 Pagans are superficial and materialistic. & Pagans have different values and priorities, and some may emphasize spiritual growth, social justice, and environmental sustainability.\\
 \hline
\end{tabular}
\caption{Example of Contrastive Synthetic Data}
\label{table:sdg}
\end{table*}

In total, we generated just under 1 million pairs of data, and filtered it down to just over 300 thousand individual samples after removing duplicates. This took around 1 day to generate given our model hosting options. We plan to openly release this dataset.

\subsection{Our Suite of Social-Bias-Detectors}
\label{sec:suite}
\texttt{social-bias} refers to the models trained on the four human curated datasets mentioned in Section \ref{sec:approach} and elaborated upon in Appendix \ref{appendix:social-bias-v0-data}. In particular, we have three such models - each trained on top of a different base model. \texttt{social-bias} is trained on top of BERT\footnote{\url{https://huggingface.co/google-bert/bert-base-uncased}}, \texttt{social-bias-distil} is trained on top of a transformer architecture \cite{trivedi2023neural} which provides most of the accuracy of a BERT-like model \cite{devlin-etal-2019-bert}, while being seven times faster on a CPU and two times faster on a GPU. Finally, \texttt{social-bias-toxigen} is trained on top of toxigen\_hatebert\footnote{\url{https://huggingface.co/tomh/toxigen_hatebert}}\cite{hartvigsen-etal-2022-toxigen}. Note that these three models were trained with only human curated datasets.

\texttt{social-bias-use-mention} refers to the models trained on the synthetic data that was generated according to Section \ref{sec:sdg}. In particular, we have four such models. We note that \texttt{social-bias-use-mention}, \texttt{social-bias-use-mention-distil}, and \texttt{social-bias-use-mention-toxigen} were trained on the entire set of unique instances from the synthetically generated data, with the base models following the same convention as above. Additionally, we trained \texttt{social-bias-use-mention-onetrial}, which saw only one iteration of synthetic data in training. We trained this model in order to provide a frame of reference for the amount of synthetic data that may be required for various levels of performance. Note that these four models have only seen synthetic data in training.

Finally, we trained \texttt{social-bias-onetrial-concat}, which contained all four human curated datasets as well as one trial of synthetic data. We trained this model as another comparison point to determine if combining human curated and synthetic data resulted in better performance.

We provide full details about each of the above models and their training data in Appendix \ref{appendix:detectors}.

\paragraph{The Cascade Approach} We also experimented with what we refer to as the cascade approach, where we utilize two models in a sequential manner. Given some text, we first identify a model, $m_{bias}$, to serve as a preliminary arbiter of whether this text contains bias or not. Then, if the output from $m_{bias}$ is the harm label, we also run the text through another model, $m_{use}$. This model determines if the text is a case of \textit{use} or \textit{mention}, as defined in Section \ref{sec:motivation}. If the output from $m_{use}$ is \textit{use}, then we assign the label of \texttt{bias} to the text. Otherwise, we assign the label of \texttt{not\_bias} - indicating that either $m_{bias}$ assigned a label of \texttt{not\_bias} or $m_{use}$ assigned a label of \textit{mention} to this text. Refer to Algorithm \ref{alg:cascade} for specific pseudo-code of the cascade approach. 

We hypothesized that the cascade would be a ``best-of-both-worlds" approach where $m_{bias}$ would do a decent job on aggregate, due to the high quality human curated datasets that it has seen. Then, to precisely combat the use-mention distinction issue, where $m_{bias}$ would incorrectly flag text that is a mention as harmful, we will run any harm-labeled instances from $m_{bias}$ through $m_{use}$, a model that is specifically trained, by way of our synthetic data, to distinguish use from mention. 

We have four cascade approaches, defined as follows:

\begin{table*}[ht]
\begin{tabular}{p{0.2\textwidth} | p{0.3\linewidth} | p{0.45\linewidth}}
 \hline
 name & $m_{bias}$ & $m_{use}$ \\ 
 \hline
 \texttt{cascade-orig} & \texttt{social-bias} & \texttt{social-bias-use-mention} \\
 \hline
 \texttt{cascade-onetrial} & \texttt{social-bias} & \texttt{social-bias-use-mention-onetrial} \\
 \hline
 \texttt{cascade-distil} & \texttt{social-bias-distil} & \texttt{social-bias-use-mention-distil} \\
 \hline
 \texttt{cascade-toxigen} & \texttt{social-bias-toxigen} & \texttt{social-bias-use-mention-toxigen} \\
 \hline
\end{tabular}
\caption{Cascade Approach Model Combinations}
\label{table:cascade}
\end{table*}

\begin{algorithm}
\caption{Cascade method}
\label{alg:cascade}
\begin{algorithmic}[1]
\State $m_{bias}$: a model which labels text as $\textit{social\_bias}$ or $\textit{not\_bias}$
\State $m_{use}$: a model which labels text as $\textit{use}$ or $\textit{mention}$
\State $data$: some dataset with a list of responses from an LLM
\State $labels$: list of labels for each response in $data$
    \Procedure{cascade}{$m_{bias}$, $m_{use}$, $data$}
    \State $labels \gets \{\}$ \Comment{initialize label list}
    \For{each response $r$ in $data$}
    \State $label_{m_{bias}} \gets m_{bias}(r)$ \Comment{run \textit{r} through $m_{bias}$}
        \If{$label_{m_{bias}}$ is $\textit{social\_bias}$} \Comment{need to run through use-mention model}
            \State $label_r \gets m_{use}(r)$\Comment{run \textit{r} through $m_{use}$}
        \Else
            \State $label_r \gets label_{m_{bias}}$ \Comment{label is $\textit{not\_bias}$ from $m_{bias}$}
        \EndIf
    \State $labels \gets labels + label_r$ \Comment{append label}
    \EndFor
    \State \Return $labels$
    \EndProcedure
\end{algorithmic}
\end{algorithm}

\section{Experiments}
\label{sec:expts}

In this section, we first describe the evaluations that we conducted and then provide empirical evidence for the cascade approach (defined in Algorithm \ref{alg:cascade}) and its utility.

\subsection{Evaluation Datasets and Baselines}
\paragraph{Collecting Evaluation Data} Due to the fact that our primary axis for evaluation is the use-mention distinction, we sought to construct an evaluation set which reflected this goal. Even though datasets for such a task are not widely available \cite{gligoric2024nlp}, we leveraged the rich literature on counter-speech and counter-narratives in the context of hate speech. In particular, similar to \cite{gligoric2024nlp}, we take the Knowledge-grounded hate countering \cite{chung-etal-2021-knowledge} and the Multi-Target CONAN \cite{fanton-2021-human} datasets. The knowledge-grounded dataset contains 195 hate speech and counter-narrative pairs covering multiple hate targets (islamophobia, misogyny, antisemitism, racism, and homophobia). The counter narratives are written by an expert who is tasked with composing a suitable counter-narrative response to a given hate speech using the corresponding knowledge as much as possible \cite{chung-etal-2021-knowledge}. The Multi-Target CONAN dataset consists of 5003 hate Speech and counter-narrative pairs covering multiple hate targets, including disabled, Jews, LGBT+, migrants, Muslims, people of color (POC), women. The dataset is constructed using a novel human-in-the-loop data collection methodology \cite{fanton-2021-human}. 

We combined both datasets and labeled each pair of \{hate speech, counter narrative\} as \{bias, not bias\}. In addition, we noted that a label of \textit{bias} most precisely meant a label of \textit{use} and similarly for \textit{not bias} and \textit{mention}. This is because counter-narratives are written such that they directly counteract the harmful hate speech \cite{gligoric2024nlp}, which results in them being excellent examples of mentions.

\paragraph{Experimental Setup} We provide two sets of evaluations below. First, in order to compare with the results from \cite{gligoric2024nlp}, we utilize the same evaluation set of 180 total examples (also taken from the above two datasets), including both hate speech and counter-narratives. Second, we combine the entirety of the Knowledge-grounded hate countering and Multi-Target CONAN datasets to arrive at an evaluation set of size $10,396$\footnote{This comes from combining 195 pairs with 5003 pairs from the Knowledge-grounded hate countering and Multi-Target CONAN datasets, respectively.}.

\paragraph{Baselines} In addition to the reported numbers with three of the GPT models from \cite{gligoric2024nlp}, we also report numbers using Llama-Guard\footnote{\url{https://huggingface.co/meta-llama/LlamaGuard-7b}} and Llama-Guard-2\footnote{\url{https://huggingface.co/meta-llama/Meta-Llama-Guard-2-8B}}. We use these two models as both a competitive baseline as well as a point of comparison due to the size and latency of these models. To reiterate, the detectors are on the order of 100M parameters, Llama-Guard models are either 7B or 8B parameters, and GPT-4 is rumored to be orders of magnitude larger \cite{openai2024gpt4}.

\section{Experimental Results}
\label{sec:results}

In this section, we provide results on the aforementioned datasets, with all of our detectors and baselines. We report three metrics of interest: false positive rate (FPR), false negative rate (FNR), and average error rate (Avg Err) - similar to \cite{gligoric2024nlp}. For clarification, the average error rate is the average of the FPR and FNR. To further contextualize these metrics, a false positive refers to incorrectly providing a label of \texttt{bias} to benign text. Moreover, because all of the examples of benign text are examples of \textit{mentions} and all the examples of harmful text are \textit{uses}, a false positive represents incorrectly flagging a mention as use. Symmetrically, a false negative refers to incorrectly flagging a use as mention. For our purposes, we note that both false positives and false negatives are important. False positives represent improper moderation by flagging benign text whereas false negatives represent missed detection by failing to flag harmful text.

We provide results below and note for all of the detectors (which are encoder-only models), generation is deterministic. For Llama-Guard models, we use a standard template\footnote{A sample notebook for inference with Llama-Guard models is available from the HuggingFace pages above} and greedy decoding.

\begin{table*}[ht]
\begin{tabular}{p{0.45\textwidth} | p{0.1\linewidth} | p{0.1\linewidth} | p{0.1\linewidth} | p{0.1\linewidth}}
 \toprule
 model & FPR & FNR & Avg Err & \# params \\ 
  \midrule
 \texttt{gpt-3.5-instruct-turbo}$^*$ & 25.56 & 13.33 & 19.44 & ? \\
 \texttt{gpt-3.5-turbo} (ChatGPT 3.5)$^*$ & 11.11 & 22.22 & 16.67 & ? \\
 \texttt{gpt-4}$^*$ & 8.89 & 20.00 & 14.44 & ? \\
 \midrule
 \texttt{Llama-Guard} & 12.22 & 20.00 & 16.11 & 7B \\
 \texttt{Llama-Guard-2} & \textit{\textbf{4.44}} & \textit{\textbf{26.67}} & \textit{\textbf{15.56}} & 8B \\
 \midrule
 \texttt{toxigen-hatebert} & 53.33 & 12.22 & 32.78 & 110M \\
 \midrule
 \texttt{social-bias} & 90.00 & 18.89 & 54.44 & 110M \\
 \texttt{social-bias-use-mention} & 34.44 & 23.33 & 28.89 & 110M \\
 \texttt{social-bias-use-mention-onetrial} & 32.22 & 37.78 & 35.00 & 110M \\
 \texttt{social-bias-onetrial-concat} & 72.22 & 20.00 & 46.11 & 110M \\
 \texttt{cascade-orig} & 32.22 & 33.33 & 32.78 & 110M x 2 \\
 \texttt{cascade-onetrial} & 26.67 & 47.78 & 37.22 & 110M x 2 \\
 \midrule
 \texttt{social-bias-distil} & 95.56 & 5.56 & 50.56 & 39M \\
 \texttt{social-bias-use-mention-distil} & 26.67 & 24.44 & 25.56 & 39M \\
 \texttt{cascade-distil} & \textbf{24.44} & \textbf{28.89} & \textbf{26.67} & 39M x 2 \\

 \bottomrule
\end{tabular}
\caption{Results on Evaluation Set with 180 examples, $^*$ denotes results taken from \cite{gligoric2024nlp}}
\label{table:paper expts}
\end{table*}

\begin{table*}[ht]
\begin{tabular}{p{0.45\textwidth} | p{0.1\linewidth} | p{0.1\linewidth} | p{0.1\linewidth} | p{0.1\linewidth}}
 \toprule
 model & FPR & FNR & Avg Err & \# params \\
 \midrule
 \texttt{Llama-Guard} & 9.27 & 5.44 & 7.36 & 7B \\
 \texttt{Llama-Guard-2} & \textit{\textbf{1.83}} & \textit{\textbf{16.62}} & \textit{\textbf{9.22}} & 8B \\
 \midrule
 \texttt{toxigen-hatebert} & 45.29 & 45.29 & 26.39 & 110M \\
 \midrule
 \texttt{social-bias} & 82.40 & 4.96 & 43.68 & 110M \\
 \texttt{social-bias-use-mention} & 36.63 & 4.10 & 20.36 & 110M \\
 \texttt{social-bias-use-mention-onetrial} & 43.31 & 9.37 & 26.34 & 110M \\
 \texttt{social-bias-onetrial-concat} & 70.55 & 3.96 & 37.25 & 110M \\
 \texttt{cascade-orig} & \textbf{32.69} & \textbf{8.31} & \textbf{19.84} & 110M x 2 \\
 \texttt{cascade-onetrial} & 35.30 & 13.49 & 24.39 & 110M x 2 \\
 \midrule
 \texttt{social-bias-distil} & 96.58 & 1.77 & 49.17 & 39M \\
 \texttt{social-bias-use-mention-distil} & 38.15 & 5.10 & 21.62 & 39M \\
 \texttt{cascade-distil} & 37.26 & 6.68 & 21.97 & 39M x 2 \\

 \bottomrule
\end{tabular}
\caption{Results on Combined Evaluation Set with 10K examples}
\label{table:full expts}
\end{table*}

\section{Discussion}
\label{sec:discussion}

\subsection{Results for 180 Examples Test Set}
First, we comment on results for the evaluation set of 180 examples from \cite{gligoric2024nlp}. Despite the small size of this test, we wanted to demonstrate performance because \cite{gligoric2024nlp} provided us with points of comparison to the GPT-family of models. In particular, we observe that our cascade models provide competitive performance with the GPT models. We make particular note of \texttt{cascade-onetrial}. Recall that the model which determines use or mention $(m_{use})$ for \texttt{cascade-onetrial} has only seen one iteration of synthetic data generation, making it extremely desirable, as the burden of training data is substantially reduced. In absolute terms, $(m_{use})$ for \texttt{cascade-onetrial} saw around 1K examples in training, whereas $(m_{use})$ for \texttt{cascade-distil} saw around 190K examples in training (refer to Appendix \ref{appendix:detectors}) for full details). Therefore, we are able to see that for this (albeit limited) set of data, the cascade approach is able to perform on par with some of the most widely used and largest models, while having a fraction of the computational cost. We do notice that both Llama-Guard models surpass even GPT-4 by a significant margin. Nevertheless, for this set of 180 examples, the cascade approach takes under 1 minute on a single A100 GPU, and is even able to be run on a CPU (taking a few hours). In comparison, access to GPT models is limited to querying via expensive external APIs, and inference with Llama-Guard models is not possible on a CPU.

\subsection{Results for Full (10K Examples) Test Set}
Next, we comment on results for the combined evaluation set of 10K examples. Here, we first point to the stellar performance from both Llama-Guard models. However, as previously mentioned, inference with Llama-Guard models is not possible on a CPU and takes several hours on an A100 GPU for a test set of this size. For our detectors, we observe that all of the detectors which have \textit{only seen human curated data} (\texttt{social-bias}, \texttt{social-bias-distil}, and \texttt{social-bias-toxigen}) have extremely high false positive rates. This implies that they perform poorly at distinguishing between use and mention, which reaffirms our initial motivations described in Section \ref{sec:motivation}. Even when we add a little bit of synthetic data and combine it with the human curated data, as in the case of \texttt{social-bias-onetrial-concat}, we see a rather high false positive rate. We also find better, but not optimal, performance in the detectors which have \textit{only seen synthetic data} (\texttt{social-bias-use-mention}, \texttt{social-bias-use-mention-distil}, and \texttt{social-bias-use-mention-toxigen}). These models tend to have lower false positive rates, but still not the best performance - implying that human curated data still contains some signal or information which is not present in the synthetic data. 

This brings us to the performance of our cascade approaches. We observe the lowest FPR among all of our detectors in \texttt{cascade-orig}. On aggregate, the cascade approaches have lower false positive rates, at the cost of slightly higher false negative rates. This is a direct consequence of the approach itself (defined in Algorithm \ref{alg:cascade}), as the method only alters labels that have already been labeled as 1 or harm (by $m_{bias}$). However, we see that the cascades are still under-performing Llama-Guard-2. We find that \texttt{cascade-orig} is around 10\% worse in terms of average error rate than Llama-Guard-2, but interestingly, the cascade approaches do better in terms of false negative rate - which suggests that they are better than Llama-Guard-2 at understanding which data points are strictly cases of hate speech or \textit{use}. Nevertheless, we observe that there is still a decent gap in the false positive rates between any cascade approach and the Llama-Guard models - implying under-performance in discerning between use and mention, with the cascade approaches having a tendency to provide a label of \texttt{bias}, leading to inaccurately labeling true \textit{mentions} as \textit{uses}. We observe that \texttt{cascade-onetrial} performs slightly worse than the best cascade, \texttt{cascade-orig}, but not too dramatically so. This is important again for the same computational efficiency reasons that were previously mentioned.

\paragraph{Significance of Low FNR} Despite the gap in false positive rates, we bring attention to the low false negative rates for our approaches. In fact, \textit{all} of the detectors and cascades have lower FNRs than Llama-Guard-2. Importantly, false negatives are directly representative of a detector's utility as a guardrail. A high false negative rate implies that the detector fails to flag text that is harmful, thereby allowing toxic content to be displayed to an end user. Hence, it is imperative to retain a lower FNR such that the detectors function as effective guardrails. We are encouraged to see as much as a 50\% decrease in FPR, at minimal expense to FNR, when comparing \texttt{social-bias} with \texttt{cascade-orig} - indicating the effectiveness of our synthetic data.

\subsection{Extended Analysis on Llama-Guard-2}
Ignoring the increased computational overhead, we further explore the differences in performance (with respect to false positives) when comparing our detectors with the Llama-Guard models. At an instance level, we explored the differences in predictive quality between Llama-Guard and our detectors. For example, the sample \textit{Did you know that children with Down syndrome, if surrounded by regularly-developed peers, receive a fundamental boost in their development? This "mix" fosters a positive learning atmosphere and increases engagement and attention in all students. This mean that children with Down syndrome would be a burden to society if isolated, not the contrary.} was incorrectly flagged as \texttt{unsafe} by Llama-Guard-2, but properly classified as \texttt{not\_bias} by our \texttt{cascade-orig} method. Another such sample is \textit{Even if all British Muslims had cancer, you are safe: you cannot catch it from anyone! Cancer cannot be spread, since it is a non contagious disease. Also, it is worth considering that illnesses are not connected to religions.} 

In both of these cases, we observe the presence of \textit{mentioning} text, which is throwing off even Llama-Guard-2. Hence, the reason for the \texttt{cascade-orig} method being able to properly classify these instances stems from the deliberate synthetic data generation and subsequent training for the use-mention distinction, but also highlights the need for continued improvement on our detectors as well to properly capture this phenomenon.

\section{Related Work}
\label{sec:related work}

\paragraph{Social Bias in Language Models} Social bias can be defined as discrimination for, or against, a person or group, or a set of ideas or beliefs, in a way that is prejudicial or unfair \cite{social_bias, Bommasani2022Trustworthy}. Numerous studies have demonstrated that generative models exhibit undesirable behavior that amplifies social bias \cite{Blodgett2020Language, Parrish2022BBQ, smith-etal-2022-im, selvam-etal-2023-tail, Dhamala2021BOLD, Nagireddy_Chiazor_Singh_Baldini_2024}. Additionally, the deficiencies of current datasets~\cite{Blodgett2021Stereotyping} and opacity of defining what constitutes social bias in language models and their measures~\cite{Blodgett2020Language, selvam-etal-2023-tail, achintalwar2024detectors} have emerged.

\paragraph{Use-Mention Distinction}\label{sec:related work-use-mention} As discussed briefly at the end of Section \ref{sec:motivation}, the \textit{use-mention distinction} can be thought of as the difference between using text for ill-intent and simply mentioning text without this malicious undertone. Although a subtle difference, this phenomenon is essential for many downstream applications. For example, a significant amount of content online falls under the umbrella of mentions, such as counter-speech, media reporting, education, and legal settings \cite{gligoric2024nlp, mun-etal-2023-beyond, kirk-etal-2022-handling, NEURIPS2022_bc218a0c, c6baca04957a44ff8cac1e822eea3a92}. Of particular use to us is the notion of counter-speech, which refers to speech produced by users of online platforms to counteract harmful speech of others \cite{gligoric2024nlp}. By definition, counter-speech is an example of \textit{mention}, and thus detection systems which are unable to distinguish use from mention risk contributing to downstream harm - such as improper removal of counter-speech. This in turn reduces opportunities to rectify false narratives and risks further censoring those already most affected by harmful language \cite{gligoric2024nlp}. Recently, there has also been work which documents the real experiences of counter-speakers and the ways in which to address existing barriers \cite{counterspeaker_perspective}. An exciting by-product of our work was our own introduction to this rich space, and we believe that the area of counter-speech will continue to provide essential help in the plight of detecting and mitigating harmful content produced by generative models.

\paragraph{Guardrails for LLMs} There is a growing set of rich literature on guardrail models and how they can be more efficient, modular, and scalable \cite{achintalwar2024detectors, rebedea-etal-2023-nemo, inan2023llama}. NeMo Guardrails \cite{rebedea-etal-2023-nemo} is an open-source toolkit for adding modular and programmable guardrails to LLM-based conversational systems. Llama-Guard \cite{inan2023llama} is an LLM-based input-output safeguard model which has a customizable taxonomy of harms. Finally, the authors in \cite{achintalwar2024detectors} go over insights from developing a host of guardrail-like models, which they refer to as detectors.

\paragraph{Synthetic Data Generation} Central to our work is the leveraging of capable large language models to generate synthetic data which can be used to train our guardrail models. In particular, we leverage taxonomy-guided data generation. Previous work such as GLAN \cite{li2024synthetic} utilizes a pre-curated taxonomy of human knowledge and capabilities as input and generates large-scale synthetic instruction data. However, as pointed out by \cite{sudalairaj2024lab}, this method relies using the proprietary GPT-4 \cite{openai2024gpt4} as the teacher model, which imposes restrictions on the downstream commercial usability of the generated data. Concretely, the terms of use of proprietary models obfuscate the viability of training commercially viable models using data that is generated from proprietary or otherwise closed models - which many times forbid using their model to improve other models \footnote{See point v under License Rights in \url{https://llama.meta.com/llama3/license/}}. Thus, similar to \cite{sudalairaj2024lab}, we utilize the open source Mixtral\footnote{\url{https://huggingface.co/mistralai/Mixtral-8x7B-Instruct-v0.1}} model to generate our data. Additionally, the taxonomy guides the generation of the data, as it enables targeted coverage around the individual leaf nodes of the taxonomy \cite{sudalairaj2024lab}.

\section{Limitations and Future Work}

\subsection{Limitations}
\label{sec:limitations}

\paragraph{Coverage of Synthetic Data} We acknowledge that our taxonomy (even the full version in Appendix \ref{appendix:taxonomy}) cannot cover all groups which are salient for social bias. This taxonomy is a point-in-time artifact, and we note that social bias dynamically evolves over time. Because social harms are the product of context-dependent classification systems with deep historical roots and are socially and morally charged, careful attention must be paid to the choices made (such as the groups included in a taxonomy) during development of the detectors \cite{achintalwar2024detectors}. Concretely, we understand that generated counter-speech may not always be reflective of real-world harm. One way to combat this, though expensive, is via human annotations with true counter-speakers \cite{counterspeaker_perspective}. Another direction is to leverage LLMs themselves to provide feedback on a given counter-speech example \cite{jones2024multiaspect}. We are experimenting with both avenues in the near future.

We also point out that closed-source LLMs tend to generate high quality data, but their proprietary nature is prohibitive for downstream use. Hence our decision to opt against use of GPT models for generation.

\subsection{Future Work}
\label{sec:future work}

\paragraph{Calibration and Confidence-Based Improvements} In order to combat overconfidence, we are considering conformal prediction approaches~\cite{vovk1999machine}. These approaches quantify uncertainty in a model's prediction by constructing \textit{predictive sets} (as opposed to singleton labels, in the case of our detectors) with guaranteed frequentist coverage probabilities. Specifically, we are looking into the regularized adaptive prediction sets approach~\cite{romano2020classification, angelopoulos2021uncertainty} which, in addition to providing coverage guarantees, produces larger (or non-singleton in the case of our detectors) prediction sets for difficult instances and smaller (singleton) sets for easier to classify examples. Then, on the instances for which our models produce these larger sets, we are exploring is the idea of a ``confident-cascade" wherein we adopt a collaborative method and offload these inputs to Llama-Guard. We conjecture this may provide a reasonable trade-off between computational efficiency and performance.

\paragraph{Novel Development} Currently, detectors are being trained in a supervised fine-tuning method with the standard binary cross-entropy loss. Given that our synthetic data is generated in a contrastive manner, we will consider training with contrastive loss \cite{khosla2021supervised}. Additionally, we are looking into directly training a multi-head detector, which eliminates the need for two models in our cascade approach.

\section{Conclusion}
\label{sec:conclusion}

In this work, we began with insights from the development of a social bias detector. Post deployment, we recounted our realization that issues arose due to the use-mention distinction. Motivated by this discovery, we described an extensible and reproducible synthetic data generation pipeline, which leverages taxonomy guided instructions in order to generate high quality labeled and contrastive data at scale. We then documented the training procedures of various models which utilized this synthetic data and introduced the cascade approach. Next, we outlined the extensive experiments performed with these models on a variety of evaluation datasets. We revealed that the cascade approach provided competitive performance on these evaluation sets, in addition to being more substantially more cost effective and compute efficient. We hope that our findings contribute to the growing body of work on building efficient and capable guardrails for large language models.

\bibliography{aaai25}

\newpage

\appendix

\section{Social Bias Detector}

As mentioned in Section \ref{sec:motivation}, we trained a social-bias-detector by fine-tuning BERT \cite{devlin-etal-2019-bert}. In particular, we took the uncased BERT model from HuggingFace\footnote{\url{https://huggingface.co/google-bert/bert-base-uncased}}. During training, we use a batch size of 16, we start with a learning rate of $1\text{e-}6$, and we train for 50 epochs, taking the best model with respect to validation f1 score. For reference, the total training time was a few hours on a single A100 GPU.

\subsection{Training Data}
\label{appendix:social-bias-v0-data}

We started with the Latent Hatred \cite{elsherief-etal-2021-latent} dataset, which is a benchmark that was specifically designed for implicit hate speech. Then, we use the 20 NewsGroups dataset \cite{misc_twenty_newsgroups_113} in order to add some out of distribution data. Note that we deliberately use this dataset in the hopes of increasing the proportion of negative (i.e., benign) labeled data in our training set. Third, we use a dataset from a work titled Identifying Implicitly Abusive Remarks about Identity Groups using a Linguistically Informed Approach \cite{wiegand-etal-2022-identifying}, as this dataset attempted to decouple harmful intent from mention of specific group identities. In fact, this concept is related to the use-mention distinction that we discuss throughout the paper. We refer to this dataset as ``Identity Groups" in Table \ref{table:initial eval results} below. Finally, we add in a subset of the CivilComments dataset \cite{civil-comments}, taking only samples which have an \texttt{identity\_attack} column value of greater than 0.5, which we believe corresponded to implicitly hateful comments.

\subsection{Evaluation}
\label{appendix:social-bias-v0-eval}

We compute evaluations with this version of our social-bias-detector on the test splits of each of the 4 datasets used in training that were mentioned above.

\begin{table*}[ht]
\begin{tabular}{p{0.2\textwidth} | p{0.1\linewidth} | p{0.1\linewidth} | p{0.1\linewidth} | p{0.1\linewidth} | p{0.1\linewidth}}
 \toprule
 test dataset & accuracy & balanced accuracy & precision & recall & f1 \\
 \midrule
 implicit-hate & 0.754 & 0.747 & 0.616 & 0.724 & 0.665 \\
 blocklisting & 0.676 & 	0.676 & - & - & - \\
 identity groups & 0.752 & 0.732 & 0.729 & 0.891 & 0.802 \\
 civil comments & 0.974 & 0.974 & 1.0 & 	0.974 & 0.987 \\
 \bottomrule
\end{tabular}
\caption{Results for \texttt{social-bias} model on evaluation sets}
\label{table:initial eval results}
\end{table*}

Some points to note:

\begin{itemize}
    \item the blocklisting data only contains benign (i.e. negatively or 0-labeled examples). Hence, precision/recall/f1 do not apply (they are trivially equal to 0)
    \item When evaluating, we predominantly focus on f1 score, in order to balance both false positives and false negatives.
\end{itemize}

\section{Taxonomy-Guided Synthetic Data Generation}
\label{appendix:taxonomy}
As mentioned in Section \ref{sec:approach}, we prompted the Mixtral mixture-of-experts model\footnote{\url{https://huggingface.co/mistralai/Mixtral-8x7B-Instruct-v0.1}}, under greedy decoding, with the instruction: \texttt{Generate a high level taxonomy for social bias that shows various groups and associated biases and stereotypes.} in order to generate a taxonomy. To be specific, we obtained a resultant taxonomy after this prompt, post-processed it, and then re-prompted the same model with the entire taxonomy passed in the context along with an instruction to \texttt{Continue the taxonomy while maintaining the structure.} We observed that one such iteration was enough to produce a viable taxonomy which can be used in the synthetic data generation process outlined in Section \ref{sec:approach}. 

We provide the full taxonomy below:
\colorlet{punct}{red!60!black}
\definecolor{background}{HTML}{EEEEEE}
\definecolor{delim}{RGB}{20,105,176}
\colorlet{numb}{magenta!60!black}
\lstdefinelanguage{json}{
    basicstyle=\small\ttfamily,
    showstringspaces=false,
    breaklines=true,
    frame=lines,
    literate=
     *{0}{{{\color{numb}0}}}{1}
      {1}{{{\color{numb}1}}}{1}
      {2}{{{\color{numb}2}}}{1}
      {3}{{{\color{numb}3}}}{1}
      {4}{{{\color{numb}4}}}{1}
      {5}{{{\color{numb}5}}}{1}
      {6}{{{\color{numb}6}}}{1}
      {7}{{{\color{numb}7}}}{1}
      {8}{{{\color{numb}8}}}{1}
      {9}{{{\color{numb}9}}}{1}
      {:}{{{\color{punct}{:}}}}{1}
      {,}{{{\color{punct}{,}}}}{1}
      {\{}{{{\color{delim}{\{}}}}{1}
      {\}}{{{\color{delim}{\}}}}}{1}
      {[}{{{\color{delim}{[}}}}{1}
      {]}{{{\color{delim}{]}}}}{1},
}
\begin{lstlisting}[language=json]
{
    "Racial and Ethnic Bias": [
        "Anti-Black bias",
        "Anti-Latino bias",
        "Anti-Asian bias",
        "Anti-Indigenous bias",
        "Anti-Middle Eastern bias",
        "Anti-White bias",
        "Anti-Romani bias",
        "Anti-Semitic bias",
        "Anti-Arab bias",
        "Anti-Indian bias",
        "Anti-Pacific Islander bias"
    ],
    "Gender Bias": [
        "Misogyny (prejudice against women)",
        "Misandry (prejudice against men)",
        "Transphobia (prejudice against transgender individuals)",
        "Non-binary erasure (prejudice against individuals who do not identify as male or female)",
        "Gender nonconformity discrimination",
        "Intersex erasure (prejudice against individuals who are born with physical characteristics that do not fit typical binary notions of male or female bodies)"
    ],
    "Religious Bias": [
        "Islamophobia (prejudice against Muslims)",
        "Anti-Semitism (prejudice against Jews)",
        "Christianophobia (prejudice against Christians)",
        "Anti-Hindu bias",
        "Anti-Buddhist bias",
        "Anti-Sikh bias",
        "Anti-Pagan bias",
        "Anti-Atheist bias",
        "Anti-agnostic bias"
    ],
    "Sexual Orientation Bias": [
        "Homophobia (prejudice against individuals who are attracted to the same gender)",
        "Biphobia (prejudice against individuals who are attracted to multiple genders)",
        "Transphobia (prejudice against transgender individuals)",
        "Asexual erasure (prejudice against individuals who do not experience sexual attraction)",
        "Panphobia (prejudice against individuals who are attracted to all genders)",
        "Polyphobia (prejudice against individuals who are attracted to multiple partners)"
    ],
    "Ableism": [
        "Disability discrimination",
        "Ableist language",
        "Sanism (prejudice against individuals with mental illness)",
        "Neurodivergent erasure (prejudice against individuals with neurological differences, such as autism or ADHD)"
    ],
    "Ageism": [
        "Prejudice against older adults",
        "Prejudice against younger adults",
        "Adultism (prejudice against children and young people)"
    ],
    "Classism": [
        "Prejudice against individuals from lower socioeconomic backgrounds",
        "Prejudice against individuals from higher socioeconomic backgrounds",
        "Economic exploitation"
    ],
    "Lookism": [
        "Prejudice against individuals based on physical appearance",
        "Colorism (prejudice against individuals with darker skin tones)",
        "Sizeism (prejudice against individuals based on their body size or shape)"
    ],
    "Xenophobia": [
        "Prejudice against individuals from other countries or cultures",
        "Cultural imperialism (the belief that one culture is superior to another and should be imposed on others)"
    ],
    "Cultural Bias": [
        "Ethnocentrism (the belief that one's own culture is superior to others)",
        "Linguisticism (prejudice against individuals who speak a different language)"
    ],
    "Political Bias": [
        "Prejudice against individuals with different political beliefs",
        "Polarization (the tendency to view those with different political beliefs as enemies rather than opponents)"
    ],
    "Environmental Bias": [
        "Prejudice against individuals who prioritize environmental concerns",
        "Climate change denial"
    ]
}
\end{lstlisting}

\section{List of Detectors}
\label{appendix:detectors}

\begin{table*}[ht]
\begin{tabular}{p{0.25\textwidth} | p{0.2\linewidth} | p{0.2\linewidth} | p{0.2\linewidth}}
 \hline
 name & base model & training data & train / val / test size \\ 
 \hline
 \texttt{social-bias} & BERT & human curated & 27113 / 8873 / 10051 \\
 \hline
 \texttt{social-bias-} \texttt{distil} & Piccolo & human curated & 27113 / 8873 / 10051 \\
 \hline
 \texttt{social-bias-} \texttt{toxigen} & toxigen\_hatebert & human curated & 27113 / 8873 / 10051 \\
 \hline
 \texttt{social-bias-} \texttt{use-mention} & BERT & full synthetic & 189669 / 31611 / 94835 \\
 \hline
 \texttt{social-bias-} \texttt{use-mention-}\texttt{distil} & Piccolo & full synthetic & 189669 / 31611 / 94835 \\
 \hline
 \texttt{social-bias-} \texttt{use-mention-} \texttt{toxigen} & toxigen\_hatebert & full synthetic & 189669 / 31611 / 94835 \\
 \hline
 \texttt{social-bias-} \texttt{use-mention-} \texttt{onetrial} & BERT & one trial synthetic & 1050 / 350 / 350 \\
 \hline
 \texttt{social-bias-} \texttt{onetrial-}\texttt{concat} & BERT & human curated, one trial synthetic & 28168 / 9225 / 10404 \\
 \hline
\end{tabular}
\caption{Suite of Social Bias Detectors}
\label{table:detectors}
\end{table*}

All models were trained using a batch size of 16, a learning rate of 0.000001, and for 50 epochs, taking the best model with respect to validation f1 score. For reference, the total training time for each model was between 30 minutes to a few hours on a single A100 GPU.

\end{document}